\documentclass[preprint,12pt]{elsarticle}

\usepackage{url}

\usepackage{breakurl}
\usepackage[breaklinks]{hyperref}
\usepackage{booktabs}
\usepackage{xcolor}
\usepackage{amsmath}
\usepackage{amssymb} 

\usepackage{graphicx} 

\usepackage{floatrow}
\usepackage{caption}
\usepackage{subcaption}

\journal{arXiv}

\makeatletter
\def\ps@pprintTitle{%
  \let\@oddhead\@empty
  \let\@evenhead\@empty
  \def\@oddfoot{\reset@font\hfil\thepage\hfil}%
  \let\@evenfoot\@oddfoot}
\makeatother

\begin{document}

\begin{frontmatter}

\title{A Remote Approach to Cashew Orchard Detection: Leveraging Active Learning with Satellite Imagery in Guinea-Bissau}


\author[fcup]{Miguel Pereira\corref{corr1}}
\cortext[corr1]{Corresponding author. Email address: up201906711@edu.fc.up.pt}
\author[inesctec]{Sofia C. Pereira}
\author[cef]{Maria J.P. Vasconcelos}
\author[cibio_biopolis]{Patrícia Guedes}
\author[cibio_biopolis]{Luke L. Powell}
\author[cmup]{João P. Pedroso}

\affiliation[fcup]{organization={Department of Computer Science, Faculty of Sciences, University of Porto, Rua do Campo Alegre, 4169-007 Porto},
            country={Portugal}}

\affiliation[inesctec]{organization={INESC-TEC, Faculty of Engineering, University of Porto, Rua Dr. Roberto Frias, 4200-465 Porto}, 
            country={Portugal}}

\affiliation[cef]{organization={
Forest Research Centre, School of Agriculture, University of Lisbon, Tapada da Ajuda, 1349-017 Lisbon}, country={Portugal}}

\affiliation[cibio_biopolis]{organization={CIBIO and BIOPOLIS, InBIO Associated Laboratory, Vairão Campus, University of Porto, Rua do Crasto, 4485-661 Vairão}, country={Portugal}}

\affiliation[cmup]{organization={CMUP and Department of Computer Science, Faculty of Sciences, University of Porto, Rua do Campo Alegre, 4169-007 Porto}, country={Portugal}}

\begin{abstract}

Cashew production is a widespread economic activity in Guinea-Bissau, as well as other countries in West Africa. However, unregulated cashew production can be directly associated with increasing regionwide deforestation rates, biodiversity losses, and a fragile economic structure. There is no nationwide database for listing or georeferencing cashew orchards, so there is a clear need to remotely map their locations. In recent years, multiple methods for detecting orchards have been developed, though they have only been applied on a regional level. This work expands regional analyses to a nationwide scale. It develops a scalable and cost-effective remote approach, based on Sentinel-2 satellite imagery, using Machine Learning techniques to detect cashew orchards automatically. Margin-based Active Learning techniques were employed to develop an optimal training set in terms of the number of points and their informativeness, leading to a cashew map with $94.0\%$ balanced accuracy obtained entirely off-site. 
We created two datasets and a 2021 cashew map with $10$m spatial resolution that are openly accessible through GitHub\footnote{GitHub link: \url{https://github.com/mrp2106/ALFICC}}.
The results demonstrate the possibility of a broader cashew orchard mapping, creating a new stepping stone for this environmental application.

\end{abstract}

\begin{highlights}
\item Cashew cultivation dominates Guinea-Bissau's land cover and economic structure
\item Mapped cashew orchards using $10\text{m}$ Sentinel-2 satellite imagery
\item Improved cashew detection performance with a new Active Learning based approach
\item Produced the first countrywide and openly accessible cashew map using 2021 data
\end{highlights}


\end{frontmatter}





\section{Introduction}
\label{sec:introduction}

\begin{figure}[!h]
    \centering
    \includegraphics[width=\linewidth]{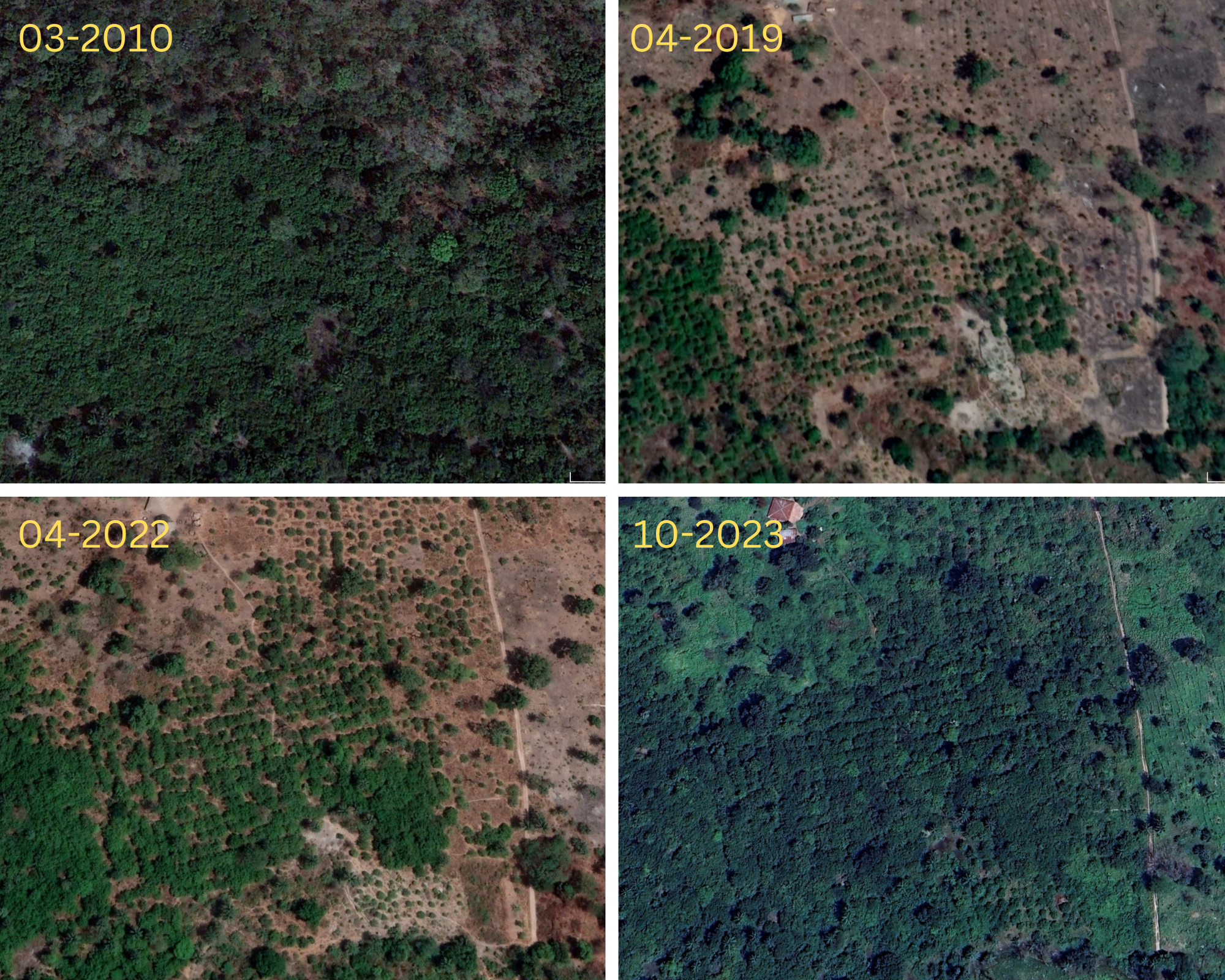}
    \caption[Transformation of forest to cashew.]{Transformation of a forested area into a cashew plantation. Note the structured row pattern in some portions of the orchard. Google Earth, Maxar Technologies/Airbus.}
    \label{fig:forest_to_cashew}
\end{figure}

Guinea-Bissau, West Africa, is suffering from an escalating environmental crisis related to increasing deforestation~\citep{h_seca2021expansion}. Although the expansion of the energy sector and illegal logging are associated with this phenomenon, deforestation in the region occurs mainly due to the conversion of forests to agricultural fields, primarily for the production of cashew, rice, and peanuts~\citep{h_seca2021expansion, c_updatedcdn_2021, powell2023eu} (Figure \ref{fig:forest_to_cashew}). Cashew nut cultivation and commerce employs many thousands of Guineans, accounts for over $90\%$ of the national exported goods, and covers between $18\%-55\%$ of the entire planted land, depending on the region~\citep{f_WFPexec_2022, a_Kone_2022, k_ensa2022cashew}. In addition to sustainability considerations, this overreliance on the cashew sector leads to a fragile single-crop economic structure. The subsistence of the country and multiple families is heavily dependent on market shifts and production values, which can be affected by pests, climate conditions, and the age of the orchard~\citep{k_ensa2022cashew, p1_undp_covid19_guinea_bissau, sierra2024insights, vasconcelos2014new, m_havik2018westafrica}. The associated habitat conversion also imposes a major biodiversity threat that could have irreversible long-term  consequences~\citep{guedes2024tipping,luis_projeto}.

Efforts to map cashew orchards have been limited due to the difficulty in differentiating between orchards and woodlands~\citep{temudo2014cashew}. However, one recent study combined Machine Learning models and Sentinel-2 images to successfully characterize cashew orchards in Cantanhez National Park, showing how much this crop has encroached into the park~\citep{34sofia_artigo}. Due to the expansion of cashew orchards across the country, more comprehensive national mapping is needed.

In this work, we propose an innovative fully remote method to detect cashew orchards and the subsequent production of a cashew map, based on the application of Machine Learning classifiers trained on remote sensing data. Land cover classes were also acquired remotely, completely eliminating the need for any on-site procedures during the project's execution.

The remote annotation process still represents a considerably time-consuming step. Motivated by this difficulty, our study explores the capacity of Active Learning (AL) techniques to optimize the class acquisition process. These techniques are applied when the learning algorithm is allowed to query an expert for labels (also called classes, in our case being land cover types) of certain samples~\citep{18tuia2011survey}. This is an iterative process, where at each step, a small batch of the currently most valuable points are passed onto the expert, which then assigns a class to each of them --- called the annotation or labeling process --- and uses them to expand the current training dataset. This way, an optimal training set with non-redundant data samples is created, which can positively impact the performance of associated ML classifier. This iterative procedure is typically repeated until either we are satisfied with the performance of the ML model, or have reached the time and/or resource budget allocated to the labeling process.


The remainder of the article is organized as follows: Section \ref{sec:related_work} contains a brief review of previous cashew monitoring and detection work, as well as the application of AL for remote data classification. Section \ref{sec:dataandmethods} describes the proposed methodology for analyzing the defined region of interest. Section \ref{sec:results} presents the results of the project's experiments, notably the AL performance and the production of a cashew map, together with a brief discussion of the obtained results. Section \ref{sec:conclusions} concludes the article.

\section{Related Work}
\label{sec:related_work}

Monitoring land use and land use changes (LULUC) is essential to understand the impact of primary sector activities on habitats and biodiversity changes. The advancements of satellite or aerial imagery, in terms of availability and spatiotemporal resolution, paired with more powerful geoprocessing tools, significantly improve the monitoring capabilities~\citep{rege2022mapping}. This has led to the expansion of cashew detection research in recent years~\citep{34sofia_artigo, rege2022mapping, 30yin2023mapping, 10nitidaemozambique}. 

The main difficulty in detecting cashew remotely comes from the identical spectral signatures of cashew and forests, as well as the typical placement of cashew trees in already forested areas, which creates a spatial mixture of both land cover types~\citep{temudo2014cashew, 34sofia_artigo,rege2022mapping}. Despite these issues, past research has shown the aptitude of ML classifiers to identify cashew orchards based on medium-high spatial resolution imagery, using either Sentinel-2 or PlanetScope satellite-acquired data~\citep{34sofia_artigo,rege2022mapping}, or sub-meter aerial images~\citep{rege2022mapping,30yin2023mapping,10nitidaemozambique}. Ensemble-based algorithms or Support Vector Machines appear as capable classifiers~\citep{34sofia_artigo,rege2022mapping}, despite their inability to work with non-tabular datasets. In contrast, deep learning algorithms are better suited for matrix-based rasters typically associated with remote sensing imagery, and can use this to extract latent spatial features that are valuable to the classification task~\citep{30yin2023mapping}. Either way, these studies agree that the addition of non-spectral variables containing either temporal or spatial-based information about each sample and its neighboring region typically helps any type of classifier in overcoming the difficulty of distinguishing spectrally similar classes, such as cashew orchards and forest woodlands, and improving overall classification performance.

The main contributions of our work to the cashew detection problem come in two fronts: the technique used and the spatial coverage. Firstly, we applied Active Learning (AL) heuristics in order to optimize the labeling process --- that is, the process of assigning labels (in our case, land cover classes) to certain selected samples (in our case, pixels). We focused on pool-based sampling techniques, which deal with large amounts of raw, unlabeled data~\citep{3thoreau2022active}. Despite the clear applicability in remote sensing classification projects with access to large amounts of raw unlabeled satellite imagery, no previous work has ever attempted to apply AL to create refined datasets for cashew mapping, which made us eager to evaluate the impact of these techniques for this specific use case. Secondly, we took advantage of the scalability aspect of a fully remote approach to perform a cashew mapping for a larger region, contrasting pre-existing work that focused on monitoring smallholder farms or protected areas~\citep{34sofia_artigo,rege2022mapping,30yin2023mapping,10nitidaemozambique}. To our knowledge, no cashew orchard map for the entire country of Guinea-Bissau has ever been created or made publicly available; thus, filling this gap was the major goal of our study. Both these fronts create an innovative approach to the study of cashew cultivation and associated effects in Guinea-Bissau, in an efficient and scalable way that is capable of keeping up with the nationwide cashew expansion and changes.


\section{Data and Methods}
\label{sec:dataandmethods}

\subsection{Study area and datasets}
\label{sec:studyarea}

\begin{figure}[h!]
    \centering
    \includegraphics[width=\linewidth]{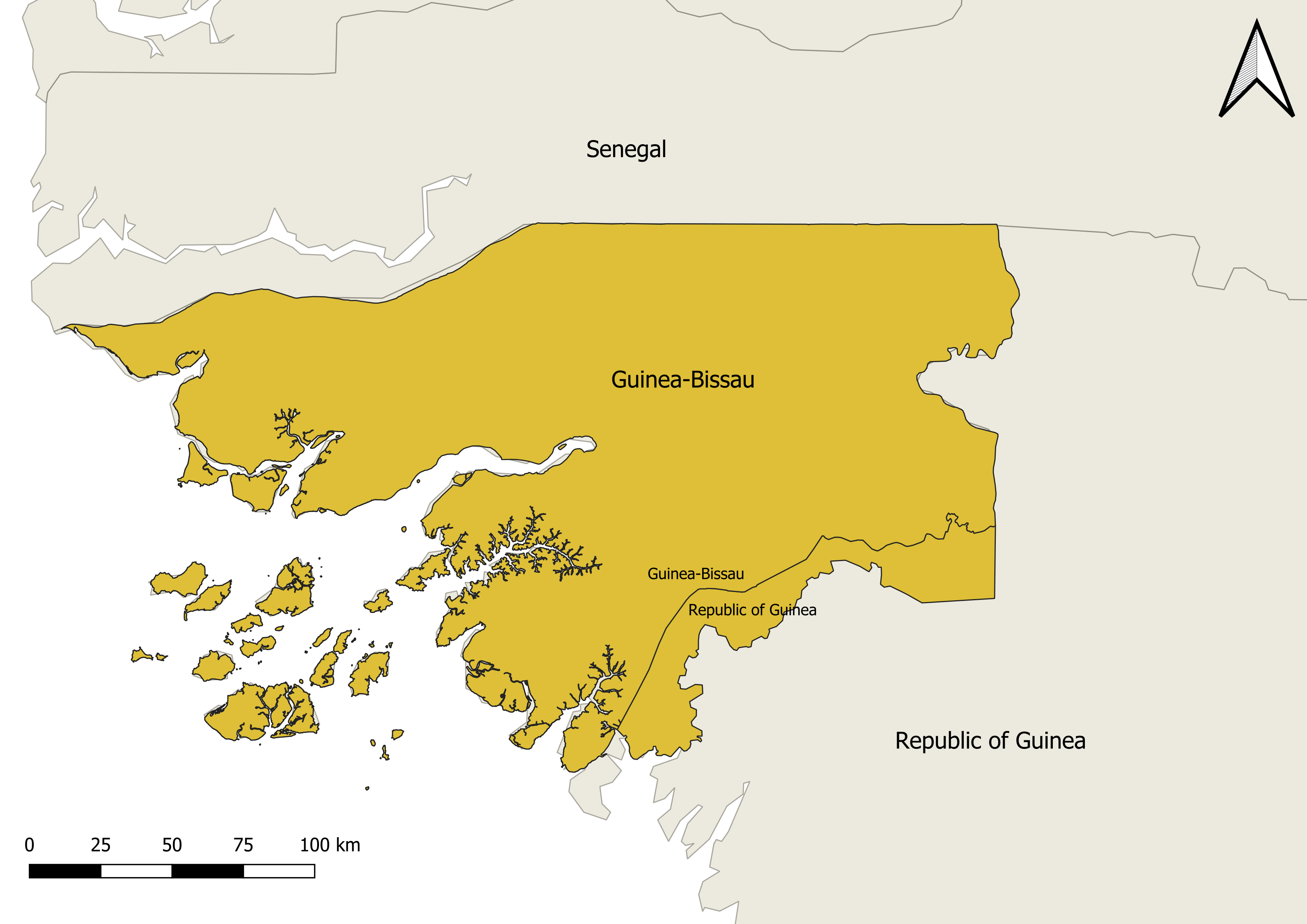}
    \caption[Study area.]{Visualization of the main study area and region of interest, the country of Guinea-Bissau, West Africa. A smaller region in the Republic of Guinea, down to the \textit{Kogon} river, is also represented. Data samples were collected in both areas. Made with QGIS~\citep{QGIS_software}.}
    \label{fig:roi_image}
\end{figure}

The project's region of interest (ROI) corresponds to the entire country of Guinea-Bissau --- mainland and islands off the western coast, as shown in Figure \ref{fig:roi_image} ---, covering an extent of $34,800\, \text{km}^2$. The region's land cover is formed by savanna woodlands, tropical and sub-tropical forests, and other open or closed-canopy landscapes~\citep{c_updatedcdn_2021}. This vegetation is marked by two seasonal behaviors: a rainy season from May to October, and a dry season from November to April. Despite its smaller area, the country holds a large plant and animal diversity and includes key biodiversity areas, which have been threatened by the habitat conversion accelerated by cashew cultivation~\citep{guedes2024tipping,luis_projeto}. Two datasets were created for the development of the study's ML and AL procedures. Samples were acquired not only in Guinea-Bissau, but also in a $2,700\, \text{km}^2$ neighboring region down to the \textit{Kogon} river in the Republic of Guinea --- also shown in Figure \ref{fig:roi_image} --- which contains very similar environmental and land cover patterns~\citep{luis_projeto}. The datasets followed different sampling techniques, namely:

\begin{itemize}
    \item \textbf{Random Sampling (RS) dataset:} corresponds to $4498$ samples (pixels) randomly selected for labeling. This heuristic represents the simplest form of expanding a training set, which is useful as a baseline comparison with other AL algorithms~\citep{2lei2021active}. Further, the unbiased and scattered selection of new samples likely ensures a diverse and representative data selection.
    \item \textbf{Margin Sampling (MS) dataset:} is comprised of $1816$ points (pixels) selected by the Margin Sampling AL heuristic. This technique requests the unlabeled samples with the smallest margin values, that is, those closest to the ML classifier's stipulated decision boundary~\citep{18tuia2011survey, 8kremer2014active}. The heuristic was chosen for its compatibility with the SVM classifier that was considered in the remainder of the project (more details in sections \ref{sec:machine_learning_workflow_and_post_processing} and \ref{sec:feature_selection_and_hyperparameter_tuning}).    
\end{itemize}

\begin{figure}[H]
     \centering
     \begin{subfigure}{0.9\textwidth}
         \centering
         \includegraphics[width=\textwidth]{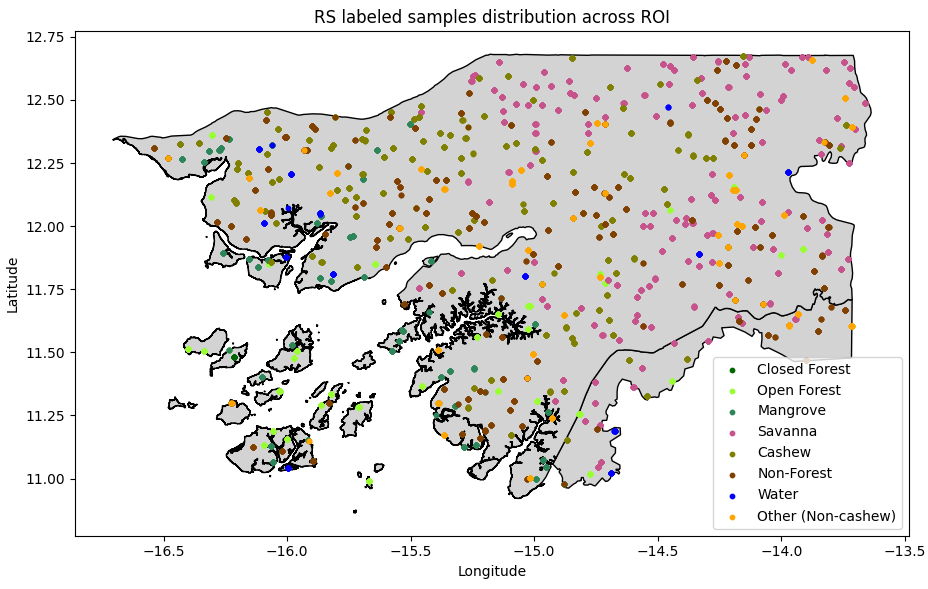}
     \end{subfigure}
     \begin{subfigure}{0.9\textwidth}
         \centering
         \includegraphics[width=\textwidth]{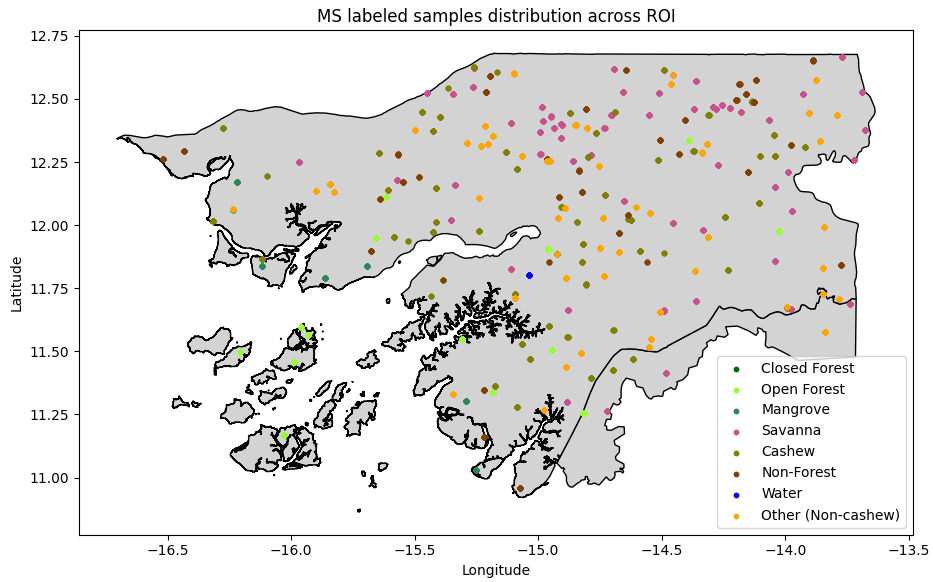}
     \end{subfigure}
    \caption[Geographical distribution of the Random Sampling samples.]{Geographical distribution of the Random Sampling (top) and Margin Sampling (bottom) datasets.}
    \label{fig:points_across_roi}
\end{figure}

Both datasets consider pixels from eight land cover classes. Our classification system follows the one utilized in the Forest Reference Emission Level report (FREL) for Guinea-Bissau~\citep{frel_gb_2019}, which was further expanded by RSeT~\citep{w_rset_home} to include more land cover types. Namely, the eight classes we considered are: Closed-canopy Forest, Open-canopy Forest, Mangrove, Savanna woodlands/ Sparse vegetation, Cashew orchards, Non-Forest, Water, and Other (Non-Cashew). Non-Forest is considered an "umbrella class", as it aggregates pixels corresponding to buildings, roads, farm fields, and empty land. Meanwhile, Other (Non-Cashew) is used to describe pixels that are definitely not cashew, but which do not fall under the scope of any other class --- for example, pixels that encompass forest and urban areas, or mangrove and water, or any other combination of multiple land cover types that thus cannot be attributed to any other individual class. These pixels can be useful in a Cashew vs. Non-Cashew classification setting, but should be discarded in any other context (more details in section \ref{sec:feature_selection_and_hyperparameter_tuning}). Figure \ref{fig:points_across_roi} shows the geographical distribution of each dataset, and Table \ref{table:pixels_per_class} presents the respective pixel counts per class.

\begin{table}[h!]
\centering
\begin{tabular}{lrr}
\toprule
 & \multicolumn{2}{c}{\textbf{Nº of pixels}} \\ 
\cmidrule(lr){2-3}
 & \textbf{Random Sampling} & \textbf{Margin Sampling} \\ 
\midrule
Closed Forest & 9 & 0 \\ 
Open Forest & 289 & 132 \\ 
Mangrove & 417 & 57 \\ 
Savanna & 1303 & 511 \\ 
Cashew & 993 & 503 \\ 
Non-Forest & 1024 & 261 \\ 
Water & 111 & 9 \\ 
Other (non-cashew) & 352 & 436 \\ 
\textbf{Total} & \textbf{4498} & \textbf{1909} \\ 
\bottomrule
\end{tabular}
\caption[Number of pixels per class in the two obtained datasets.]{Number of pixels per class in the two obtained datasets.}
\label{table:pixels_per_class}
\end{table}

\subsection{Satellite image processing}
\label{sec:satellite_image_processing}

\begin{figure}[h!]
    \centering
    \includegraphics[width=\linewidth]{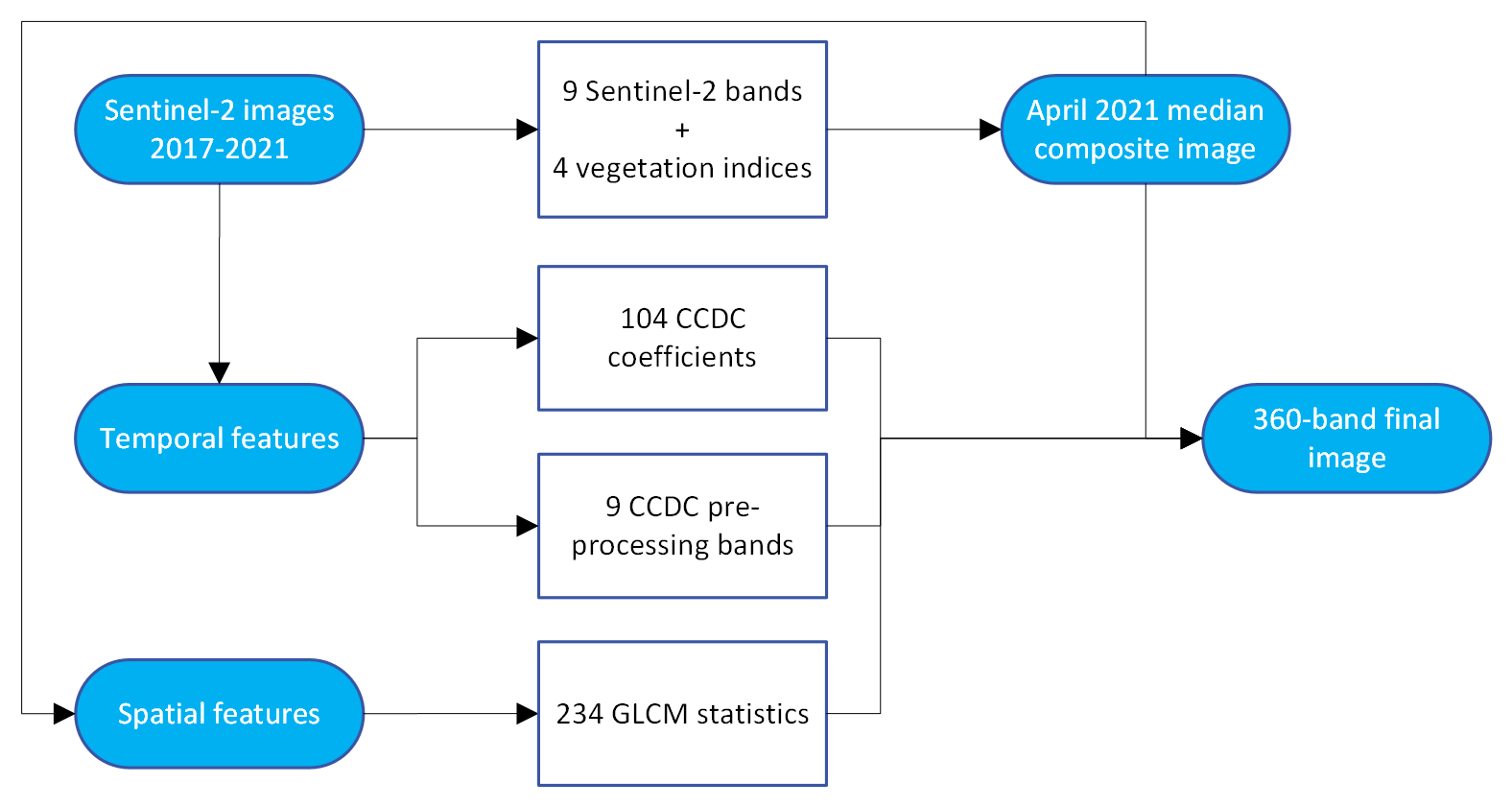}
    \caption[Satellite image processing and feature acquisition flowchart.]{Satellite image processing and feature acquisition flowchart.}
    \label{fig:fluxogram_satellite_processing_feature_acquisition}
\end{figure}

Remote sensing imagery from Copernicus Sentinel-2 (S2) was used as the base satellite data~\citep{o_s2mission}. This mission is composed of two orbiting satellites phased at $180^{\circ}$, producing a new image with almost global coverage with a five-day revisiting time. The satellites collect radiance measurements across $13$ discrete bands in the electromagnetic spectrum's visible, near-infrared, and shortwave-infrared regions, with spatial resolutions of either $10$m, $20$m, or $60$m. In our work, every non-$10$m band was resampled to a $10$m spatial resolution, which matches the resolution of every considered additional band --- thus, the produced cashew map also presents a $10$m spatial resolution. Satellite image processing and feature extraction were performed in Google Earth Engine (GEE), a cloud-based platform that combines open-access imagery with distributed computational resources for advanced processing~\citep{gee_article_sofia_article}. Figure \ref{fig:fluxogram_satellite_processing_feature_acquisition} presents a flowchart of this procedure.

An image collection was considered using the S2 images from March 2017 up to April 2021. These data were collected and stored in tiles --- $109.8\text{km} \times 109.8\text{km}$ segments of the Earth's surface ---, which can be assembled to form a spatially continuous image for the entire region of interest through a process called mosaicking~\citep{34sofia_artigo}. From the resulting $13$-band raster collection, training features were extracted to develop the ML classifiers. They are categorized and shown in the flowchart across three different branches according to the nature of the information they represent:

\begin{itemize}
    \item \textbf{Spectral features:} Four Sentinel-2 bands were discarded from the analysis: bands B1, B9, and B10 are typically ignored in surface-level applications, as they account for atmospheric effects; also, band B8A was dropped due to its spectral similarity and overlap with band B8. Four bands related to different vegetation indices were added to the base image collection, corresponding to the NDVI, NBR, EVI, and EVI2 indices. 
    Spectral overlap between different land cover types is a typical concern during the classification procedure, noticeably between cashew orchards and certain types of forest woodlands. This overlap is not the same at every time of the year and is typically smaller by the end of the dry season, which spans from November to April~\citep{34sofia_artigo}. Therefore, we combined all the base S2 images from April 2021 and extracted the median image with the $13$ spectral bands through a process called compositing --- using an aggregation function to merge spatially overlapping imagery into a single image.
    \item \textbf{Temporal features:} Temporal information can highlight other distinctive characteristics of land cover types, such as the seasonal variability of savanna woodlands or the stability of a closed-canopy forest, thus enhancing the predictive capacity of an associated classifier. For this purpose, we applied the Continuous Change Detection and Classification (CCDC) algorithm to the S2 2017-2021 image collection~\citep{q_zhu2014continuous}. 
    The procedure works in two stages: initially, a change detection algorithm identifies breaks in the typical behavior of each pixel --- which can be associated with a land cover change ---, partitioning a pixel's time series into segments; secondly, a harmonic regression is applied to each of the resulting segments for each pixel. The coefficients of the harmonic regression can then be extracted and used as temporal predictors for the classification procedure. For each pixel, we extracted $104$ coefficients of the time segment overlapping with the $30^{th}$ of April 2021, plus nine other bands that were also added during the preprocessing phase of GEE's CCDC implementation regarding Elevation, Rainfall, and other geographic-related features, creating a $113$-band image.
    \item \textbf{Spatial features:} The most common type of manually extracted features for multispectral image classification regard spatial information~\citep{34sofia_artigo}. The similarity or contrast with neighboring pixels can be used to detect, for example, the sparse pattern of savanna woodlands in the dry season, as well as the structured tree pattern that concerns certain cashew orchards, such as the one shown in Figure \ref{fig:forest_to_cashew}. 
    We utilized the Gray-Level Co-Occurrence Matrix (GLCM) algorithm to calculate spatial statistics~\citep{glcm_medium}. This method works with gray-scale intensities --- valued between 0 and 255 --- to describe a central pixel's neighboring region, using metrics like the sum average, entropy, local contrast, and dissimilarity between pairs of pixels. The GEE implementation calculates 18 statistics per input band and thus was only applied to the April 2021 median composite. This resulted in a $234$-band image, which was merged with the spectral April 2021 median composite and the CCDC image to produce the final image asset, containing $360$ features for every pixel in the region of interest.
\end{itemize}

\subsection{Active Learning and label acquisition}
\label{sec:active_learning_and_label_acquisition}

\begin{figure}[h!]
    \centering
    \makebox[\textwidth]{%
        \includegraphics[width=1.3\linewidth]{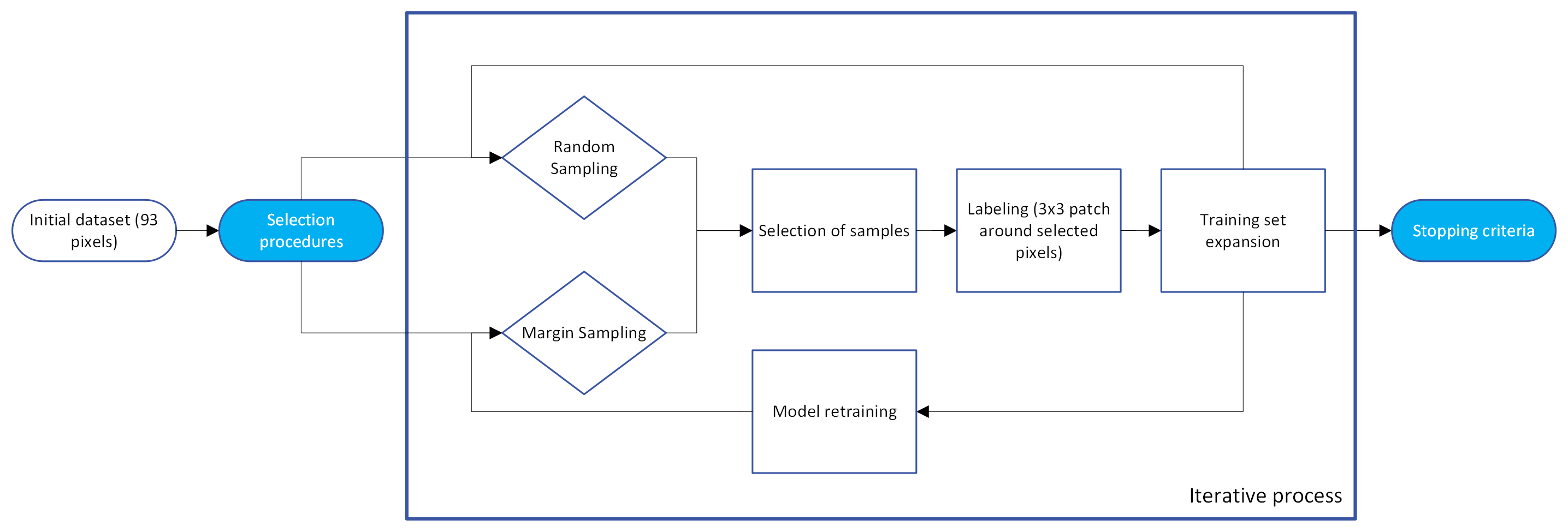}
    }
    \caption[Active Learning and dataset creation schema.]{Schema of the Active Learning and dataset creation process.}
    \label{fig:al_flowchart}
\end{figure}

Figure \ref{fig:al_flowchart} shows a flowchart of the sampling techniques and AL procedures that led to the creation of the Random Sampling and Margin Sampling datasets. An initial training set common to both sampling heuristics was stipulated before any other data acquisition process because the evolution of the AL iterative procedure is directly influenced by its training data. We separated $11$ randomly acquired polygons --- $93$ pixels --- for this purpose, having at least one polygon of each of the eight annotated classes to ensure that every classifier does not have to discover new classes during the iterative labeling process, and two polygons for Savanna, Cashew, Non-Forest, and Other (Non-Cashew).

The labeling process was accompanied by RSeT researchers~\citep{w_rset_home} with previous experience from other remote or on-site projects in the area. Google Earth Pro's Historical Imagery tool (Landsat/Copernicus/Maxar Technologies/Airbus) was used to view the terrain and assign classes according to each pixel's land cover in April 2021, at the end of the dry season. In each Random and Margin Sampling iteration, after the algorithm requested pixels for labeling, the entire $3 \times 3$ patch of the immediate neighboring pixels was also stored and given to the annotator (the first author) for labeling, which sped up the annotation process. Often, cases were too difficult to classify with confidence, mostly due to the fuzzy nature of land cover over such a large and diverse area (multiple pixels contained mixed behaviors and were thus not suited to be categorized into one single class), or the lack of high-resolution imagery in Google Earth Pro to assess the land cover type in 2021. In such cases, the samples were discarded.

The final step of the AL process is the stopping criteria, which stipulate when the iterations and dataset creation end. The literature typically suggests stopping the acquisition process either when the associated model performance reaches satisfactory levels or when the time and labeling budgets have been exhausted. However, we also considered the possibility of stopping if a significant portion of the samples requested by the heuristic became overwhelmingly hard to annotate and/or had to be discarded. Thus, we decided that the AL procedure should be terminated when one of the three aforementioned criteria was reached.

\subsection{Machine Learning workflow and post-processing}
\label{sec:machine_learning_workflow_and_post_processing}

All Machine Learning and image classification procedures were implemented in Python~\citep{python_reference_manual}. This includes the stipulation of a test set, feature selection, and hyperparameter tuning for model selection, with every step being performed after the creation of the Random Sampling (RS) dataset, but before the acquisition of Margin Sampling (MS) data, since this heuristic's iterative selection of points is dependent on these choices. A train-test split of the RS data was performed using a grouped stratified sampling strategy, which deals with data representation in each set with stratified sampling; the approach keeps all the pixels from the same patch together in either training or testing to avoid spatial autocorrelation affecting the test scores~\citep{34sofia_artigo}. The dataset was split $50-50$ to ensure a highly representative test set. Meanwhile, the MS dataset was only used for model training - every MS result was also evaluated on the RS test set to ensure the comparability with the RS results.

Due to the high number of bands in the raster described in section \ref{sec:satellite_image_processing} ($360$), feature selection was performed on the RS training set, following two different procedures: \textit{scikit-learn}'s~\citep{sklearn_software} recursive feature elimination algorithm, which iteratively removes the least prominent features that have little to no impact on the model's predictive performance; a feature-importance based approach, where we utilized scikit-learn's Random Forests \textit{feature\_importances} attribute to get an estimate of each variable's usefulness with a value between 0 and 1, and only kept those with a score above the threshold of $\frac{1}{360}$, which represents the average importance in a random classifier scenario (where every feature would have the same impact on the model's performance).


Hyperparameter tuning followed the bayesian optimization strategies of the \textit{hyperopt} Python library~\citep{hyperopt_software}. Two types of classifiers were considered for optimization: Support Vector Machines (SVMs) and Random Forests (RFs). SVMs had already shown the highest cashew detection performance in previous research~\citep{34sofia_artigo}. RFs have been considered in the aforementioned research as well, and despite not presenting the best results, they are one of the few ML algorithms implemented in GEE's native Machine Learning tools, which can facilitate the transition between the satellite image processing and ML sections. The hyperparameter search was performed optimizing mean F1-Cashew on $5$-fold cross-validation in the RS training set for both models.


During hyperparameter tuning, we explored two different classification scenarios: firstly, we discarded the samples from the eighth class (Other non-cashew) and worked with 7-class classifiers; after, we considered binary classifiers to work in a Cashew vs. Non-Cashew classification setting (this time including the eighth class into the broader Non-Cashew category). We performed hyperparameter tuning separately for each case to observe if a noticeable performance difference could arise from this label change. As discussed in section \ref{sec:feature_selection_and_hyperparameter_tuning}, the binary SVM classifier outperforms the 7-class model, and as such was the considered model for the AL section.

The final classifier of the MS experiments, trained on the full MS dataset, was used to create a prediction-based land cover map for the entire region of interest. A final post-processing step was performed by applying a sieve filter to the land cover map. This filter removes raster polygons that are smaller than a pre-defined threshold and replaces them with the pixel value (class) of the largest neighboring polygon, thus smoothing the land cover map. The optimal threshold value will be discussed in section \ref{sec:post_processing_and_final_land_cover_map}. 



\section{Results and Discussion}
\label{sec:results}



\subsection{Feature selection and hyperparameter tuning}
\label{sec:feature_selection_and_hyperparameter_tuning}


During the feature selection process, the recursive feature elimination algorithm selected $250$ of the initial $360$ variables, while the above-average importance process kept only $67$ features. Figure \ref{fig:feature_selection_f1_test} in \ref{sec:appendix_feature_selection} shows that the latter does not translate to any performance drop in comparison to the recursive feature elimination procedure or the usage of the total $360$ features, even presenting a slightly higher median score than the other two. Hence, we only kept the $67$ features selected by the above-average procedure for the remaining ML procedures and the training of the Active Learning classifiers.



The details of the hyperparameter tuning section, including the search spaces for each classifier type and the architecture of the best classifier in both 7-class and 2-class scenarios, are available in \ref{sec:appendix_feature_selection}. Table \ref{table:hyperparameter_tuning_scores} shows the test scores of the best 7-class algorithm, a Random Forest, and the 2-class model, an SVM. Both balanced accuracy and Cashew F1 scores are higher for the binary SVM classifier. Naturally, the binary prediction task is less complex: collapsing the original Non-Cashew land cover classes into just a single category erases minority classes - such as Closed Forest and Water (Table \ref{table:pixels_per_class}) - which typically drag the metrics down. 



\begin{table}[h!]
\centering
\begin{tabular}{lrr}
\cmidrule[\heavyrulewidth]{2-3}
 & \textbf{Balanced Accuracy} & \textbf{F1-score Cashew} \\ 
\midrule
\textbf{Best 7-class model (RF)} & 0.803 & 0.814 \\ 
\textbf{Best 2-class model (SVM)} & 0.905 & 0.843 \\ 
\bottomrule
\end{tabular}
\caption[Test results for the best classifiers after hyperparameter tuning.]{RS test results for the best 7-class and 2-class classifiers. Hyperparameter search was made for maximizing mean F1-Cashew on $5$-fold cross-validation
in the RS training~set.}
\label{table:hyperparameter_tuning_scores}
\end{table}

Due to our main goal of detecting cashew cultivation, we have decided to work with the binary classification approach and the best 2-class model from this point onward. Because the best binary classifier is an SVM, we opted to use a margin-based heuristic, specifically MS, as the AL heuristic to build the dataset.

\subsection{Active Learning results}
\label{sec:active_learning_results}

\begin{figure}[h!]
    \centering
    \includegraphics[width=\linewidth]{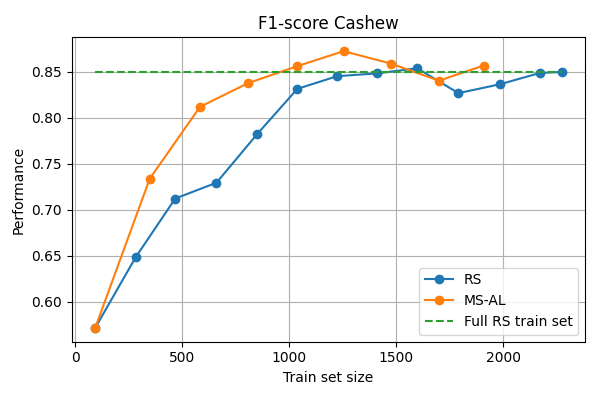}
    \vspace{-0.8cm}
    \caption[Active Learning iterative results.]{Cashew F1-score evolution during the Active Learning procedure}
    \label{fig:ms_rs_al_iterations_results_right_half}
\end{figure}

After optimizing a classifier on the Random Sampling train set and selecting an initial training set for the Active Learning process, we advanced to the acquisition of newly labeled samples to form the Margin Sampling dataset. We also simulated an AL evolution on the RS training data by iteratively selecting larger portions of this dataset, to serve as a baseline comparison. The results are presented in Figure \ref{fig:ms_rs_al_iterations_results_right_half}.

Both selection procedures lead to improvements of the associated ML classifiers --- that is, the performance of the ML model increases as we add new labeled pixels to the training set in each RS or MS iteration, until it stagnates. However, MS does so more efficiently, with steeper increases during the early iterations: just after two AL steps, the classifier obtained an F1-score of $81.2\%$ using MS, in comparison to the $71.2\%$ score using randomly selected data. Peak performance is also higher with MS labeled training data, reaching $87.2\%$ Cashew F1-score with $1254$ training samples --- in comparison, RS's peak is $85.4\%$ F1-score with $1599$ samples.

The results of the final iterations show a degree of oscillation. This could mean the classifier is reaching the maximum predictive performance, though we cannot claim this with certainty, considering this behavior is only observed for a small number of final iterations. However, during the MS annotation task, the performance decay after the fifth iteration's peak was noticeably accompanied by the request of more intricate samples and the increase in sample rejection rates (the proportion of requested samples by the AL heuristic for which the annotators could not attribute a land cover class), as shown in Figure \ref{fig:ms_percentage_discarded_points_per_iteration} in \ref{sec:appendix_al_experiments_and_ms_rejection_rates}. Thus, even if we hadn't already reached the performance cap, we would have stopped the annotation process due to the increased difficulty in labeling new pixels with confidence, which could steeply increase the risk of mislabeling and data intrusion, whilst also adding little new informative samples to the train set.


Therefore, the AL experiments finished with a Balanced Accuracy of $92.2\%$ and F1-score of $85.7\%$ after nine iterations, with 1909 samples. Random Sampling finalizes with $90.9\%$ Balanced Accuracy and $85.0\%$ F1-score, using 2274 training samples. The results show the power of Active Learning in the context of multi-spectral image classification and cashew detection, as it allows for a more efficient training sample selection, which helps improve a classifier's maximum predictive performance and also leads to better results with fewer labeled data. 

\subsection{Post-processing and final land cover map}
\label{sec:post_processing_and_final_land_cover_map}

With the final Margin Sampling classifier (the SVM trained on the full MS dataset), we obtained a cashew land cover map from the predictions for every pixel in the ROI. In order to apply the sieve filter to smoothen the map, we first needed to stipulate the minimum polygon size $p$. Since this is a user-defined threshold, we searched for the optimal value in a grid. As shown in Figure \ref{fig:sieve_search_results_cut} in \ref{sec:appendix_sieve_filter_and_communities_overlay}, the best results are obtained with $p = 350$, which leads to a Cashew F1-score of $89.5\%$ and balanced accuracy equal to $94.0\%$, an improvement of both metrics over the unfiltered results, as interpreted in the figure.

\begin{table}[h!]
\centering
\begin{tabular}{cccc}
\textbf{Cashew (ha)} & \textbf{Non-Cashew (ha)} & \textbf{Cashew (\%)} & \textbf{Non-Cashew (\%)} \\
\midrule
\multicolumn{1}{r}{902,149} & \multicolumn{1}{r}{2,575,932} & \multicolumn{1}{r}{25.9} & \multicolumn{1}{r}{74.1} \\
\bottomrule
\end{tabular}
\caption[Area estimates and percentage of Cashew and Non-Cashew in Guinea-Bissau.]{Area estimates and percentage of Cashew and Non-Cashew in Guinea-Bissau.}
\label{table:cashew_percentages_in_land_cover_map}
\end{table}

\begin{figure}[h!]
    \centering
    \includegraphics[width=\linewidth]{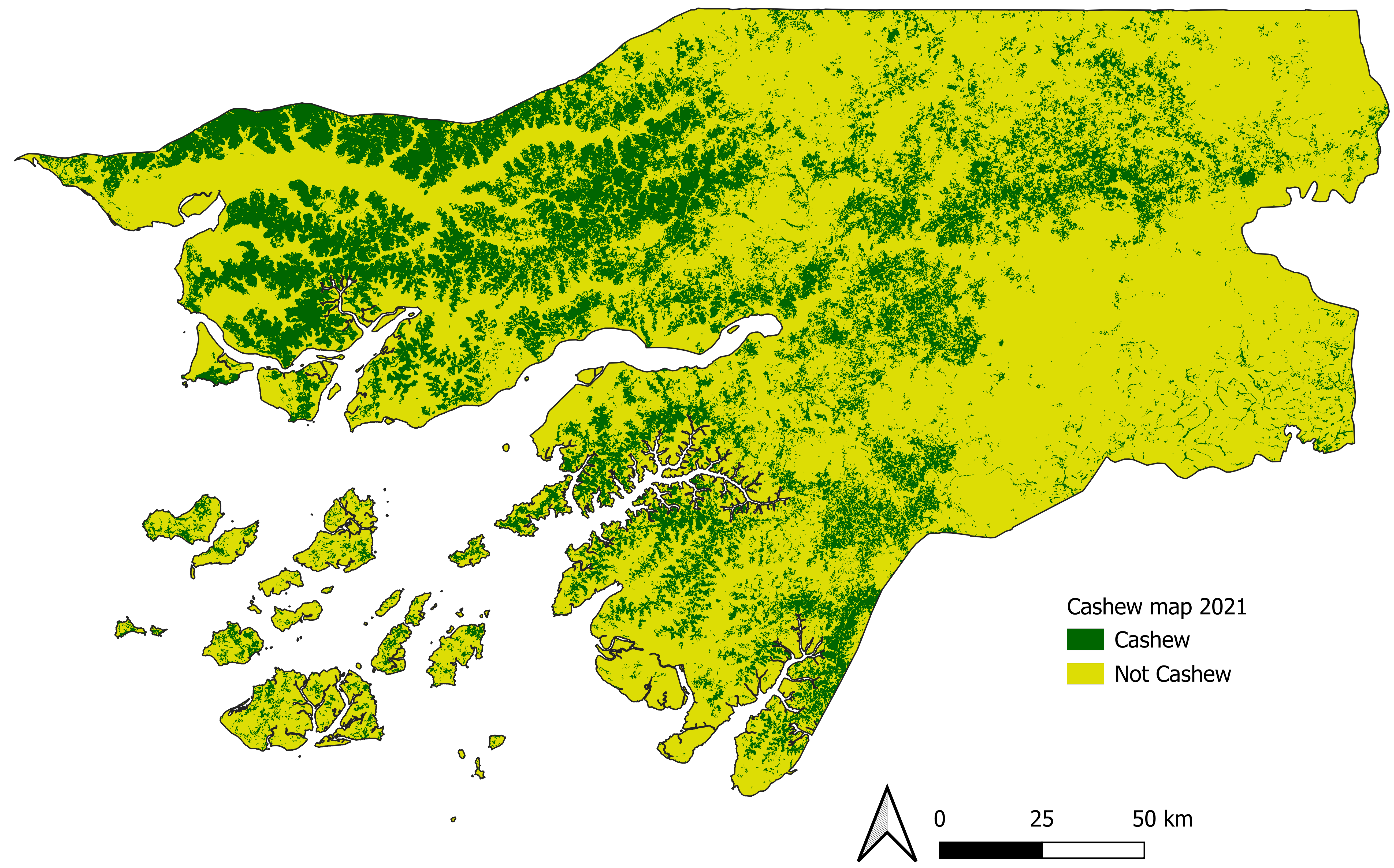}
    \caption[Final 2021 Land cover map.]{Final 2021 Land cover map. Obtained with a post-processing $350p$ sieve filter applied to the SVM predictions map.}
    \label{fig:ms_al_sieve_best_boundaries}
\end{figure}

The filtered 2021 land cover map for cashew monitoring, based on SVM predictions, is shown in Figure \ref{fig:ms_al_sieve_best_boundaries}. In total, $25.9\%$ of the country's area is identified as Cashew, as presented in Table \ref{table:cashew_percentages_in_land_cover_map}, almost equal to the percentage of terrestrial protected areas~\citep{34sofia_artigo}. The northwest quadrant of the country contains the most cashew, followed by the southwest and northeast. Cashew plantations are scarcer in the southeast. These results constitute the basis for an accurate assessment of the state of the cashew sector in the country, which then allows the implementation of appropriate measures to control its prevalence.

Further insights can be drawn from this study, and particularly from the cashew land cover map. For example, the relationship between cashew cultivation and human activity becomes clear in the overlay presented in Figure \ref{fig:ms_al_best_communities} in \ref{sec:appendix_sieve_filter_and_communities_overlay}, as it highlights the geographical relationship between cashew orchards and communities. Another example is presented in Figure \ref{fig:ms_al_best_protectedareas}, which shows the spatial intersection between the cashew map and Guinea-Bissau's terrestrial protected areas (PAs). This example highlights an opposite relation to that of the communities overlay, as the PAs typically intersect regions with fewer cashew plantations (aside from Cantanhez National Park, whose case was studied in~\citep{34sofia_artigo}). This visual interpretation already hints at the correlation of cashew cultivation with both phenomena, though further analysis is needed to draw out causal effects from these interactions, if they exist.

\section{Conclusions}
\label{sec:conclusions}




This study is the first attempt at an openly accessible cashew mapping for Guinea-Bissau, West Africa. Through the combined use of an SVM classifier paired with an Active Learning procedure that developed a minimal but optimized training dataset, we were able to produce a fully remote and cost-effective approach to detect cashew orchards. Margin Sampling proved to be an efficient sample selection algorithm, resulting in $89.5\%$ Cashew f1-score and $94.0\%$ balanced accuracy test results after a sieve filter was applied to the predictions map. 

Cashew cultivation has a deep environmental, economic, and humanitarian impact in many West African countries, including Guinea-Bissau~\citep{guedes2024tipping}. This effect is exacerbated by the unregulated expansion of plantations, which in turn complicates the accurate estimation of cultivation areas and production values. Recently, researchers have identified mapping and monitoring cashew orchards as a priority for a sustainable future in West Africa~\citep{guedes2024tipping,34sofia_artigo}. Our work directly contributes to this by serving as the basis for a deeper assessment of the state of the cashew sector in the area, which can lead to more adequate and effective action focused on the country's sustainable development. Further, our approach can be applied to other Western African countries where cashew orchard expansion is a threat to biodiversity, facilitating an accurate measure of the area this crop currently occupies. 

Further developments can be made to expand the approach presented in this study. Cashew mapping accuracy might increase with the usage of more advanced patch-based classifiers~\citep{22tong2020land}, or through the creation of a new training dataset based on a different AL heuristic. It would also be valuable to map land cover across years to develop a change detection system that could then be used to monitor the expansion of cashew orchards in Guinea-Bissau and beyond.



\newpage

\appendix

\section{Feature selection and hyperparameter tuning}
\label{sec:appendix_feature_selection}

\begin{figure}[H]
    \centering
    \includegraphics[width=0.9\linewidth]{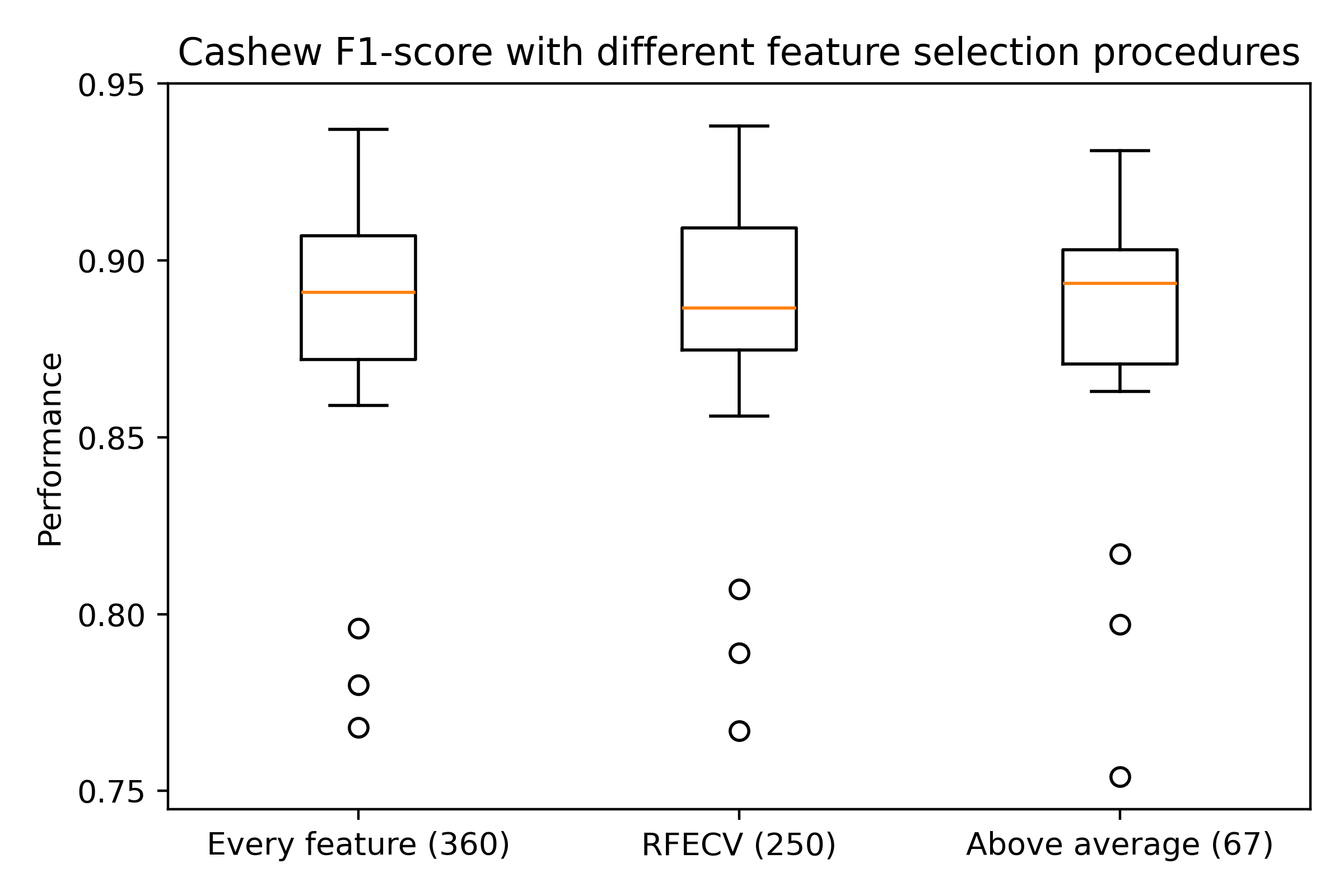}
    \caption[Feature selection results]{Performance with three different variable sets: every feature (360), the selection through recursive feature elimination RFECV (250), and features with above average importance value (67). Results were obtained with stratified grouped $10$-fold on the Random Sampling training set.}
    \label{fig:feature_selection_f1_test}
\end{figure}

Figure \ref{fig:feature_selection_f1_test} compares the results obtained using every feature with two variable selection procedures: recursive feature elimination (RFECV) and selecting features with an importance score above the $1/360$ threshold that corresponds to the mean (average) importance. Considering the performance similarity between the different scenarios, the boxplots demonstrate that we can reduce the dimensionality of the feature space without sacrificing model performance. This justifies our choice of working with the smallest feature space composed of the $67$ variables with an above-average importance score.

Regarding the hyperparameter tuning process, Table \ref{table:hp_search_spaces} shows the considered hyperparameters for RFs and SVMs, and the respective search spaces. A separate search was performed for 7-class and 2-class classification paradigms, with the best models from each scenario being detailed in Table \ref{table:hp_best_models}. The AL results that followed were obtained using the best binary SVM classifier due to its higher cashew prediction scores, as shown in Table \ref{table:hyperparameter_tuning_scores}.

\begin{table}[h!]
\centering
\begin{tabular}{ll}
\toprule
\multicolumn{1}{c}{\textbf{Hyperparameters}} & \multicolumn{1}{c}{\textbf{Search range}} \\
\cmidrule{1-2}
\multicolumn{1}{c}{\textbf{Random Forests}} &  \\
Number of estimators & $50-450$, with steps of 10 \\
Maximum depth & $1-12$ \\
Max features per split & $3-70$, with steps of 3 \\
\cmidrule{1-2}
\multicolumn{1}{c}{\textbf{Support Vector Machines}} &  \\
C (regularization parameter) & $10^{-3} - 10^1$ \\
Kernel & Linear, \textit{poly}, or \textit{rbf} \\
Gamma & $10^{-10} - 10^1$ (for \textit{rbf} or \textit{poly} kernel) \\
Degree & 2, 3, or 4 (for \textit{poly} kernel) \\
\bottomrule
\end{tabular}
\caption[Hyperparameter search space for RFs and SVMs.]{Hyperparameter search space for Random Forests and Support Vector Machines. The search space was equivalent for the 7-class and 2-class classification paradigms.}
\label{table:hp_search_spaces}
\end{table}

\begin{table}[H]
\begin{tabular}{lllll}
\toprule
\multicolumn{2}{c}{\textbf{2 class}} &  & \multicolumn{2}{c}{\textbf{7 class}} \\
\midrule
\textbf{Best model} & \textbf{SVM} &  & \textbf{Best model} & \textbf{RF} \\
C & $0.0117$ &  & Number of estimators & 73 \\
Kernel & poly &  & Maximum depth & 10 \\
Gamma & $0.6439$ &  & Max features per split & 2 \\
Degree & 4 &  &  & \\
\bottomrule
\end{tabular}
\caption[Best classifiers after hyperparameter tuning.]{Best classifiers after hyperparameter tuning, for both 2-class and 7-class classification scenarios.}
\label{table:hp_best_models}
\end{table}

\section{AL experiments and MS rejection rates}
\label{sec:appendix_al_experiments_and_ms_rejection_rates}

\begin{figure}[H]
    \centering
    \includegraphics[width=0.7\linewidth]{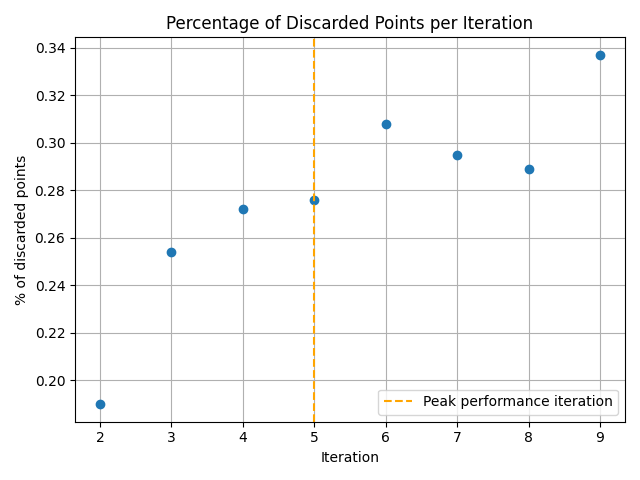}
    \vspace{-0.2cm}
    \caption[Percentage of discarded pixels per MS iteration.]{Percentage of the number of requested pixels per MS iteration that could not be labeled and were thus discarded.}
    \label{fig:ms_percentage_discarded_points_per_iteration}
\end{figure}

Figure \ref{fig:ms_percentage_discarded_points_per_iteration} highlights the increase in labeling difficulty after the peak performance iteration, for the MS labeling process. After the peak, the heuristic started requesting more and more difficult pixels, often with a mixture of land cover classes, which translated in an increase in the number of discarded samples. This was especially evident during the final iteration, where over a third of the pixels requested by the model could not be labeled.

\section{Sieve filter and communities overlay}
\label{sec:appendix_sieve_filter_and_communities_overlay}

\begin{figure}[H]
    \centering
    \includegraphics[width=0.7\linewidth]{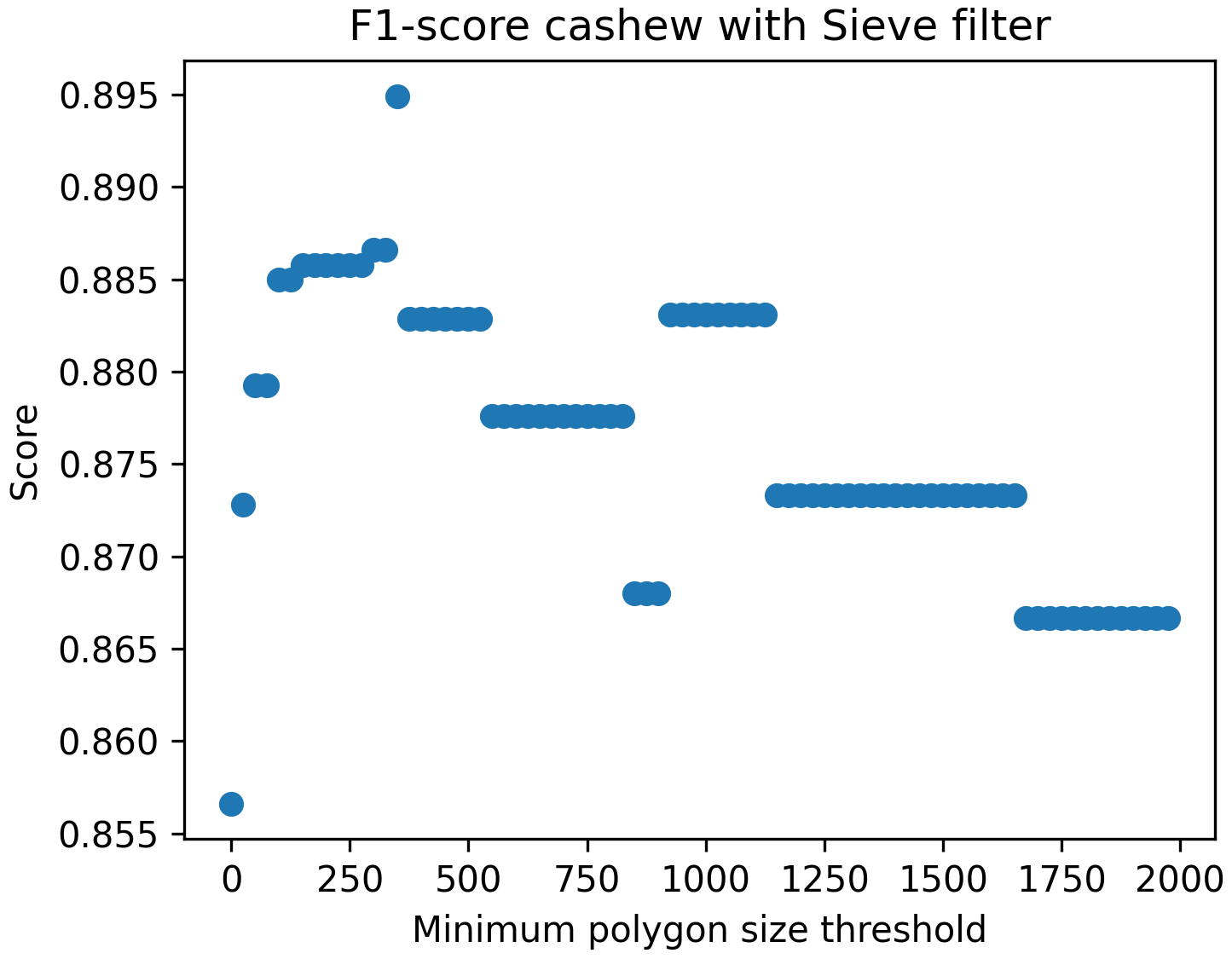}
    \vspace{-0.4cm}
    \caption[Sieve filter with different polygon threshold test scores.]{Test scores obtained after applying a sieve filter with different polygon size threshold $p$ parameter values. The first point, with minimum polygon size threshold $p$ equal to $0$, represents the test scores obtained with no filter applied.}
    \label{fig:sieve_search_results_cut}
\end{figure}

Figure \ref{fig:sieve_search_results_cut} shows how the map scores vary with the sieve filter polygon size threshold. The first data point represents the scenario where no filter is applied to the map. The results show that the application of such a filter is beneficial, since the test scores are always higher when applying a sieve filter, no matter the value of the polygon size threshold.

\begin{figure}[H]
    \centering
    \includegraphics[width=0.88\linewidth]{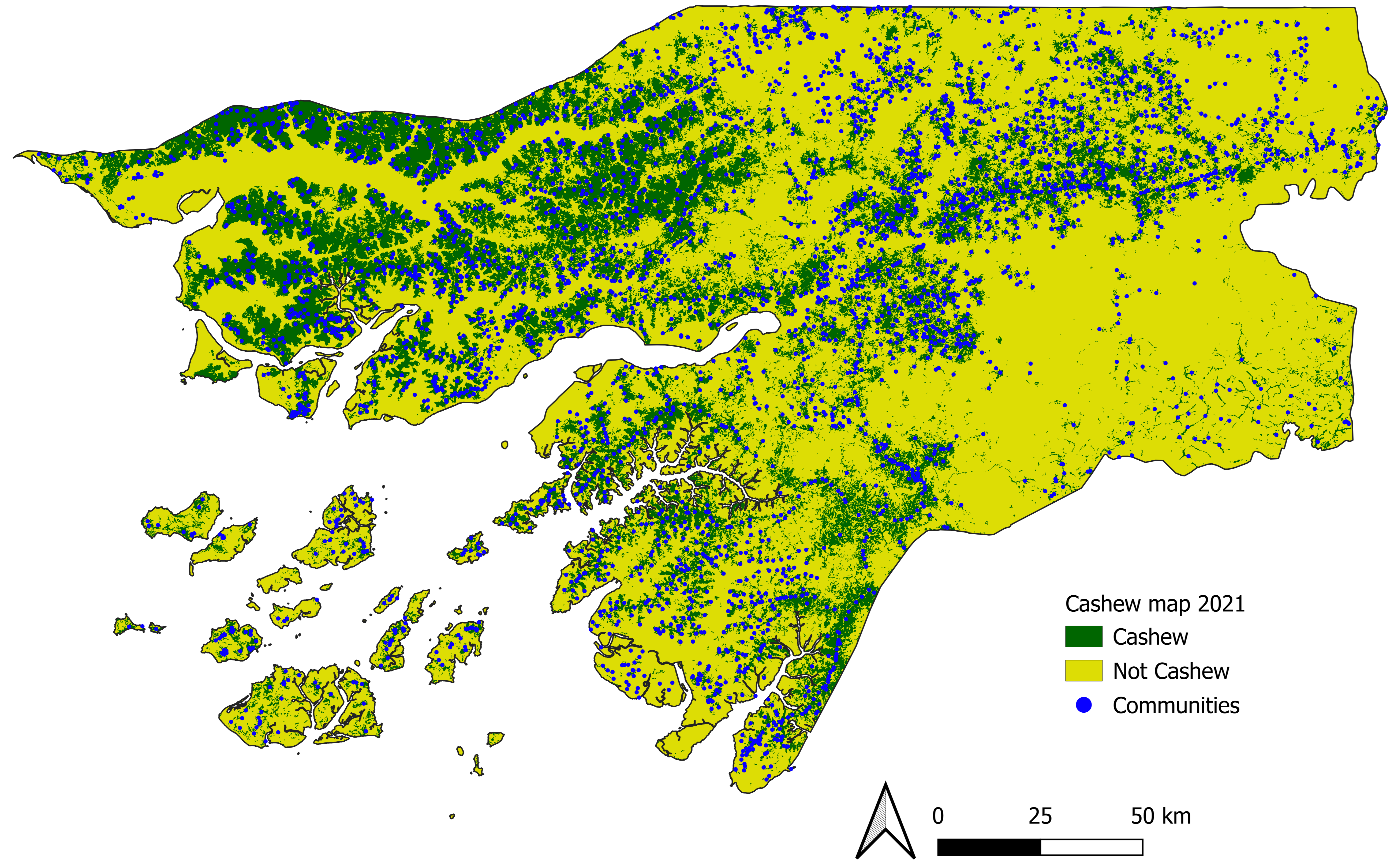}
    \caption[Land cover map with an overlay of Guinea-Bissau communities.]{Land cover map with an overlay of Guinea-Bissau communities according to the 2017 national census.~\citep{w_rset_home}}
    \label{fig:ms_al_best_communities}
\end{figure}

\begin{figure}[H]
    \centering
    \includegraphics[width=0.88\linewidth]{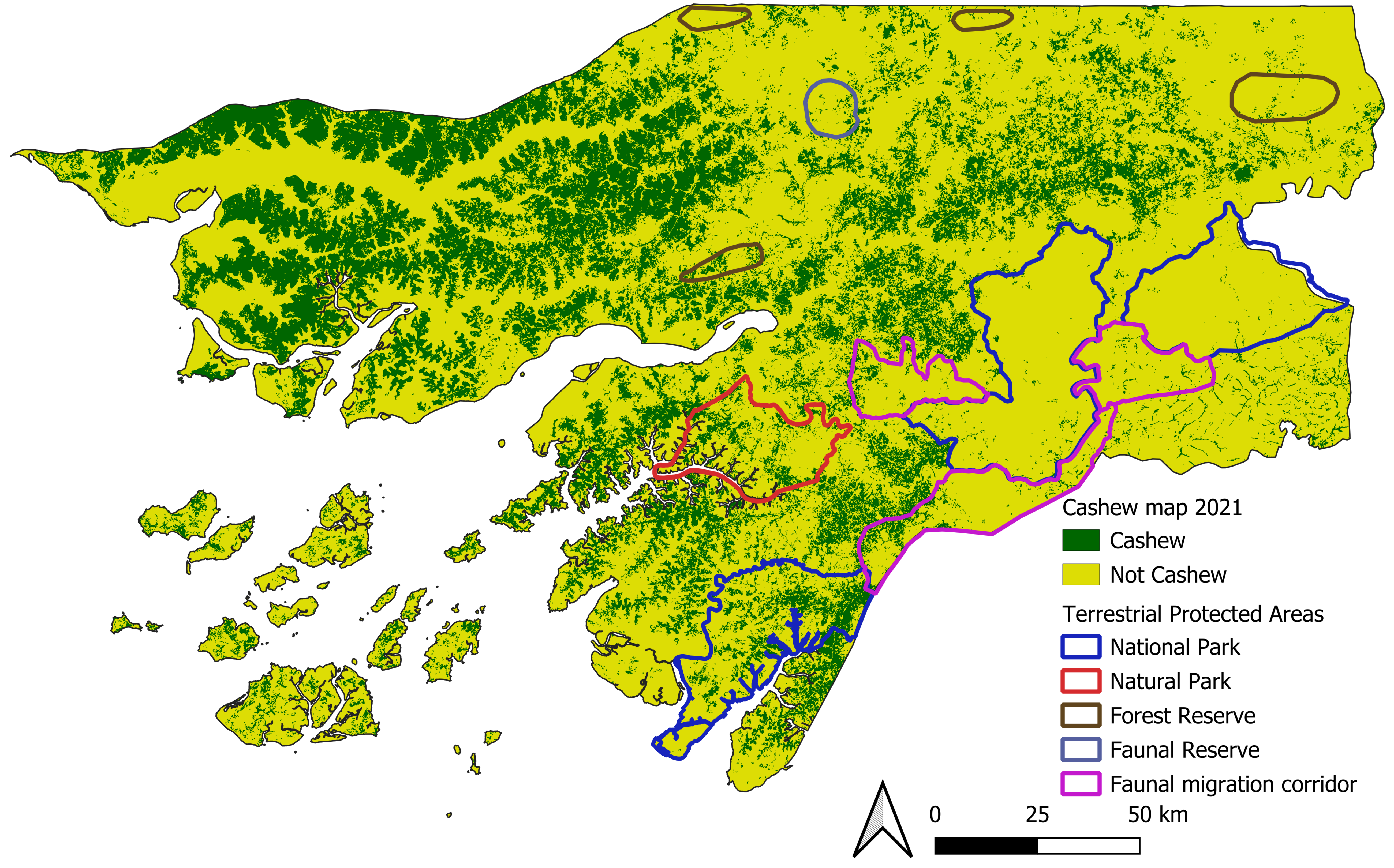}
    \caption[Land cover map with an overlay of Guinea-Bissau Protected Areas.]{Land cover map with an overlay of Guinea-Bissau Terrestrial and Inland Waters Protected Areas.~\citep{protected_planet_PAs}}
    \label{fig:ms_al_best_protectedareas}
\end{figure}

Figure \ref{fig:ms_al_best_communities} shows the spatial overlap between the largest cashew predominant regions and existing urban locations, overlaying Guinea-Bissau's 2017 census data on a land cover map processed with a $350p$ sieve filter. This evidence is clearer in the Northern part of the country. In addition, Figure \ref{fig:ms_al_best_protectedareas} presents the overlay with Guinea-Bissau's terrestrial protected areas (PAs), showing the opposite tendency: most protected areas appear in southeastern regions, with little cashew orchards (aside from Cantanhez National Park). The northwest quadrant, which shows the highest orchard density, also coincides with the absence of any terrestrial PAs.

\section*{Acknowledgements}

We want to thank Luís Palma for his valuable insights and input about the results and project applicability, as well as RSeT~\citep{w_rset_home} for providing us with supporting technical documentation and datasets that were crucial for the early project development.

This work was co-funded by the European Union’s Horizon 2020 research and innovation programme under grant agreement No 854248 and by the project TROPIBIO NORTE-01-0145-FEDER-000046, supported by Norte Portugal Regional Operational Programme (NORTE2020), under the PORTUGAL 2020 Partnership Agreement, through the European Regional Development Fund (ERDF).

The author João Pedro Pedroso was partially supported by CMUP, member of LASI, which is financed by national funds through FCT – Fundação para a Ciência e a Tecnologia, I.P., under the projects with reference \\ UIDB/00144/2020 and UIDP/00144/2020.

\bibliographystyle{bibliography_style} 
\bibliography{Bibliography}

\end{document}